\pdfoutput=1

\documentclass[11pt]{article}

\usepackage[final]{acl}

\usepackage{times}
\usepackage{latexsym}
\usepackage{placeins}
\usepackage{float}
\usepackage{tabularx}
\usepackage{hyperref}

\usepackage[T1]{fontenc}

\usepackage[utf8]{inputenc}
\usepackage{caption}
\usepackage{amsmath}
\usepackage{amssymb}
\usepackage{booktabs}
\usepackage{microtype}
\usepackage{algorithm}
\usepackage{algorithmicx}
\usepackage{algpseudocode}
\usepackage{enumitem} 
\usepackage{tikz}
\usepackage{multirow}
\usepackage{inconsolata}

\usepackage{graphicx}
\usepackage{svg}

\usepackage{float}

\title{MoE-I$^2$: Compressing Mixture of Experts Models through Inter-Expert Pruning and Intra-Expert Low-Rank Decomposition}

\author{
 \textbf{Cheng Yang\textsuperscript{1*}},
 \textbf{Yang Sui\textsuperscript{1,2*}},
 \textbf{Jinqi Xiao\textsuperscript{1}},
 \textbf{Lingyi Huang\textsuperscript{1}},
 \textbf{Yu Gong\textsuperscript{1}},
 \textbf{Yuanlin Duan\textsuperscript{1}},
 \\
 \textbf{Wenqi Jia\textsuperscript{3}},
 \textbf{Miao Yin \textsuperscript{3}},
 \textbf{Yu Cheng\textsuperscript{4}},
 \textbf{Bo Yuan\textsuperscript{1}}
\\
 \textsuperscript{1}Rutgers University,
 \textsuperscript{2}Rice University,
 \\
 \textsuperscript{3}The University of Texas at Arlington,
 \textsuperscript{4}The Chinese University of Hong Kong
\\
 \small{
 \texttt{cheng.yang@rutgers.edu, yang.sui@rice.edu, bo.yuan@soe.rutgers.edu}
 }
}

\begin{document}
\maketitle
\begin{abstract}
The emergence of Mixture of Experts (MoE) LLMs has significantly advanced the development of language models. Compared to traditional LLMs, MoE LLMs outperform traditional LLMs by achieving higher performance with considerably fewer activated parameters. Despite this efficiency, their enormous parameter size still leads to high deployment costs. In this paper, we introduce a two-stage compression method tailored for MoE to reduce the model size and decrease the computational cost. First, in the inter-expert pruning stage, we analyze the importance of each layer and propose the Layer-wise Genetic Search and Block-wise KT-Reception Field with the non-uniform pruning ratio to prune the individual expert. Second, in the intra-expert decomposition stage, we apply the low-rank decomposition to further compress the parameters within the remaining experts. Extensive experiments on Qwen1.5-MoE-A2.7B, DeepSeek-V2-Lite, and Mixtral-8$\times$7B demonstrate that our proposed methods can both reduce the model size and enhance inference efficiency while maintaining performance in various zero-shot tasks. The code will be available at \url{https://github.com/xiaochengsky/MoEI-2.git}

\end{abstract}

\def\thefootnote{*}\footnotetext{Equal Contribution.}
\section{Introduction}
\label{sec:introduction}






Large Language Models (LLMs) have recently demonstrated remarkable language understanding and generation proficiency, excelling in complex tasks~\cite{OpenAI, LLama, wu-etal-2020-de}. However, deploying these models presents substantial challenges due to their significant storage and computational demands. To overcome these issues, the Mixture-of-Experts (MoE) LLM has been proposed~\cite{Mixtral}, which activates only a subset of its parameters during training and inference. For instance, with a smaller model size, the Mixtral-8$\times$7B model with a total of 47B parameters surpasses the performance of dense Transformer models like LLaMA-2-70B~\cite{LLama2}. Additionally, Qwen1.5-MoE-A2.7B~\cite{Qwen} demonstrates highly competitive performance compared to other 7B models, and the recently introduced DeepSeekv2 MoE~\cite{DeekSeekv2} achieves performance levels comparable to GPT-4, demonstrating the powerful capabilities of MoE models.

MoE models have garnered significant attention recently due to their ability to dynamically select subsets of parameters for each input, enabling efficient handling of diverse tasks. Despite their potential, a notable challenge with MoE models is that they are still burdened by substantial parameter size and computation cost. For example, Mixtral-8$\times$7B~\cite{Mixtral} not only has 47B parameters but also activates 13B parameters during inference. While this architecture allows for scalability and flexibility, it also introduces complexities and huge memory in deployment and inference, particularly when considering resource constraints and efficiency. Consequently, decreasing and maintaining these large-scale models remains a critical area of research.

Model compression techniques, such as pruning, knowledge distillation, and quantization, have been utilized to slim the model size. ~\cite{EEP} proposed to reduce the parameter count of MoE models by expert pruning, but it does not reduce the parameters during inference efficiently. ~\cite{merge} merges several experts into one and applies the low-rank decomposition to further reduce the model size. Although this approach achieves a good compression ratio and performance, it requires calibration and fine-tuning for each downstream task individually, which is not suitable for large-scale LLMs, and time costs are very high. Several works ~\cite{Zhou2021, Sun2023, SparseGPT} focus on unstructured sparsity to decrease the parameters of models while maintaining high performance. However, unstructured pruning struggles to achieve practical acceleration, decrease inference, and save storage without a specific design for hardware and libraries.

To solve these problems, we start by analyzing parameter redundancy in the MoE model from multiple levels. First, since identifying redundant experts using brute-force search ~\cite{EEP} is infeasible in practice, it is necessary to design efficient methods to reduce the time complexity. Second, we aim to compress as many experts as possible while ensuring that the model maintains its zero-shot performance, rather than being limited to handling a single down-stream task~\cite{merge}. Finally, our method can adapt to any MoE model, particularly those with a large number of experts and diverse structures, and automatically identifies a suitable compression strategy for each type of MoE model without the need for manual settings.

In this paper, we propose a novel end-to-end framework for MoE models, MoE-I$^2$, for the task-agnostic compression of the MoE models. To our knowledge, MoE-I$^2$ is the first end-to-end framework designed task-agnostic for structured compression of MoE LLMs. Our contributions are summarized as follows:
\begin{itemize}
\item We introduce a two-stage MoE compression framework for expert slimming that considers both inter-expert and intra-expert relationships.  
\item In the inter-expert pruning stage, we analyze the importance of each MoE layer and propose a non-uniform pruning ratio for each layer. Then, we find that previous MoE pruning methods lead to high time complexity and local optima. To address these issues, we introduce a layer-wise genetic search to reduce time complexity and a block-wise combination strategy to approximate a global optimum better.
\item In the intra-expert decomposition stage, we measure the importance of each expert and assign non-uniform ranks accordingly. Subsequently, we apply a low-rank decomposition to further compress the parameters within each expert in a fine-grained manner.
\item We conduct extensive experiments with MoE models, including Qwen1.5-MoE-A2.7B (14.3B), DeepSeek-V2-Lite (16B), and Mixtral-8$\times$7B (47B), across various nine datasets to assess both the generation quality and the zero-shot classification performance, demonstrating the effectiveness of our proposed MoE-I$^2$ framework.
\end{itemize}

\section{Related Works}
\label{sec:related-work}

\subsection{Mixture-of-Experts LLMs}
MoE-LLMs have gained significant attention in recent years due to their ability to scale efficiently while maintaining high performance. MoE models divide the network into several experts and dynamically select a subset of these experts for each input, which reduces computational overhead and enhances scalability. ~\cite{Shazeer2017} introduced the MoE model in their work on the Sparsely-Gated Mixture-of-Experts Layer, and ~\cite{GShard2020} further advanced the MoE architecture by demonstrating its scalability to trillions of parameters while retaining manageable computation costs by distributing the experts across multiple devices. With the recent advancements in decoder-only architecture\cite{LLama}, MoE models built on this structure have become increasingly popular ~\cite{Mixtral}. In this paper, we focus on how to build an end-to-end framework to solve post-training expert pruning and decomposition for MoE LLMs to decrease computation and storage.

\subsection{Compression on MoE LLMs}
Recent advancements in large language models have underscored the need to reduce parameter sizes and latency~\cite{llmpruner}. Compression techniques for language models include network pruning~\cite{networkpruning2021}, knowledge distillation~\cite{kd2019,kd2020}, quantization~\cite{quan2022}, decomposition~\cite{FWSVD2022,ASVD2023,SVDLLM2024}, and methods like early exit~\cite{DeeBERT2020}. Building on these techniques, pruning, and sparsity is crucial for MoE models, which often have up to 95\% of parameters dedicated to experts. Pruning MoE models involves removing less important experts or neurons to reduce the number of active parameters during inference. For example, ~\cite{kim2021scalable} retains the most activated experts to enhance machine translation MoE models, while ~\cite{Koishekenov2022} introduces gate statistics-based pruning during decoding. Although effective, these methods are mostly confined to linguistic models in machine translation. ~\cite{chen2022} dropping-while-training approach progressively removes non-essential experts for specific tasks, tested on Switch Transformers ~\cite{Switch2021}. The merge-compression ~\cite{merge} method and EPP ~\cite{EEP} approach, which is similar to ours, consider pruning and skipping in MoE models but face challenges in reducing computational costs. Given a pruned or sparse model, finetuning aims to restore performance on original tasks. Recent studies on LLMs ~\cite{Sun2023, llmpruner} focus on pruning linear layers, but these methods often fail to reduce computing costs without specialized hardware or libraries. Efficient post-finetuning expert pruning and sparsity methods for task-agnostic MoE LLMs remain underexplored. This gap highlights the need for advanced techniques to effectively balance pruning and sparsity while maintaining or enhancing performance across various tasks.

\section{Method}
\label{sec:method}

\begin{figure*}[htbp] 
\vspace{2mm}
\centering 
\includegraphics[width=\textwidth]{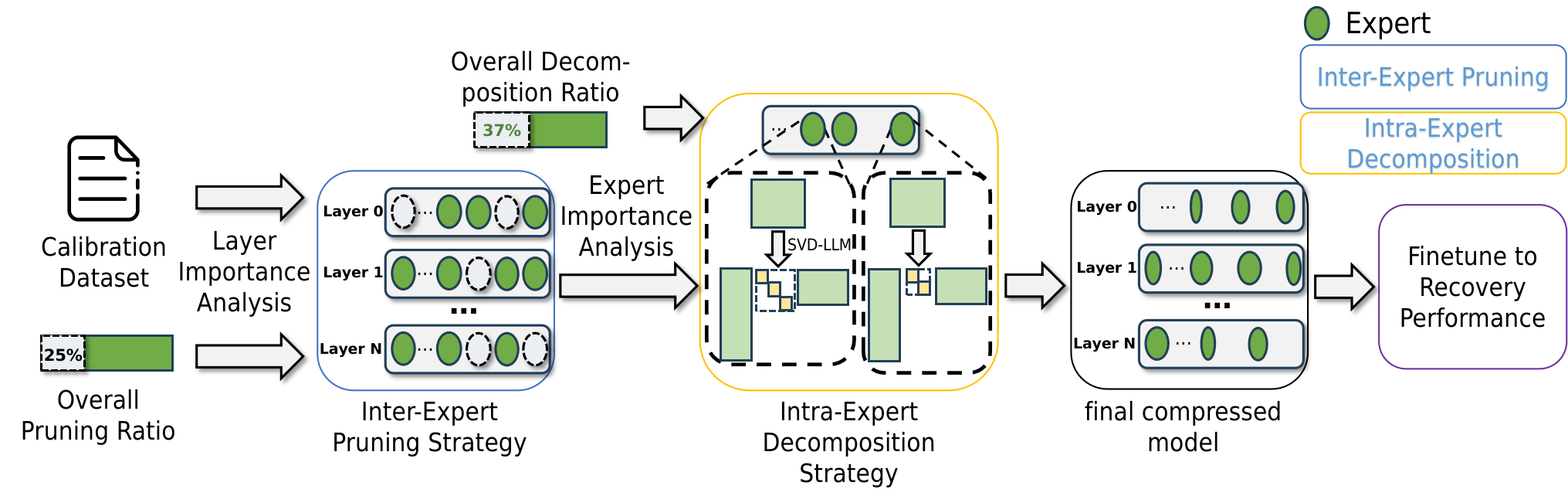}
\caption{The three-stage pipeline of MoE-I$^2$. The first stage (left) represents \textit{Inter-Expert Pruning}, where MoE-I$^2$ conducts the \textit{Layer Importance Analysis} on the target MoE model. By using a predefined overall pruning rate, it determines varying pruning ratios of different layers. Subsequently, the unimportant experts in MoE are determined by Layer-wise Genetic Search and Block-wise KT-Reception Field. The MoE is pruned accordingly. The second stage (middle) represents \textit{Intra-Expert Analysis}. Similarly, MoE-I$^2$ automatically performs \textit{Expert Importance Analysis} on the pruned model and using a predefined overall decomposition rate, applies varying ranks and low-rank decomposition to different experts, resulting in a final compressed model. The third stage (right) shows that we fine-tuned the compressed MoE model to recover performance.}
\label{fig:pipeline} 
\vspace{2mm}
\end{figure*}

In this section, we introduce the details of our proposed framework, MoE-I$^2$, which consists of three stages: Inter-Expert Pruning stage (Sec.~\ref{sec:inter}), Intra-Expert Decomposition stage (Sec.~\ref{sec:decomposition}), and fine-tuning stage (Sec.~\ref{sec:finetuning}). The overall pipeline is shown in Figure~\ref{fig:pipeline}.

\subsection{Inter-Expert Pruning}
\label{sec:inter}

In this stage, our goal is to prune individual unimportant experts to reduce the parameter size and computational cost. It raises two crucial questions: (1) \textit{Given an overall pruning ratio, how many experts should be pruned in each layer?} (2) \textit{How to determine which experts to prune?} 

\subsubsection{Layer Importance Analysis}
\label{sec:layer_analysis}

To answer the first question, we start by analyzing the importance of each layer. The layer importance of $i$-th layer, denoted by $I_i$, is defined as the average loss degradation by removing individual experts within this layer. Specifically, to calculate $I_i$ in the $i$-th layer, we first calculate the expert importance. We consecutively pruning $j$-th expert in the $i$-th layer, denoted by $e_{i,j}$, where $j = 1, 2, \cdots, M_i$. The $M_i$ represents the total number of experts in the $i$-th layer. Next, each pruned model predicts the next token with the calibration samples. The expert importance of $e_{i,j}$ is calculated as:
\vspace{-2mm}
\begin{equation}
I_{i,j} = \sum_{B} \mathcal{L}(\mathcal{X}, \{\mathbf{E}_{i}\} \setminus \{e_{i,j}\})
\vspace{-2mm}
\label{eq:expert}
\end{equation}
where $\{\mathbf{E}_{i}\} = \{e_{i, 1}, e_{i, 2}, \cdots, e_{i, M_i}\}$ denotes the set of all experts in $i$-th layer. $\mathcal{X}$ represents the calibration dataset, and $B$ denotes the batche size. $\mathcal{L}$ denotes the output of the MoE model under the condition that the $j$-th expert in the $i$-th layer is removed. Once we have determined the importance score of the $j$-th expert in the $i$-th layer, the overall importance score of $i$-th layer is defined as $I_i = \sum_{j=1}^{E_i} I_{i,j}$. Given the overall pruning rate, we normalize the layer importance to obtain the pruning rate for each layer.

\begin{figure*}[htbp]
\vspace{2mm}
    \centering
    \begin{minipage}[b]{0.32\textwidth}
    \includegraphics[width=\textwidth]{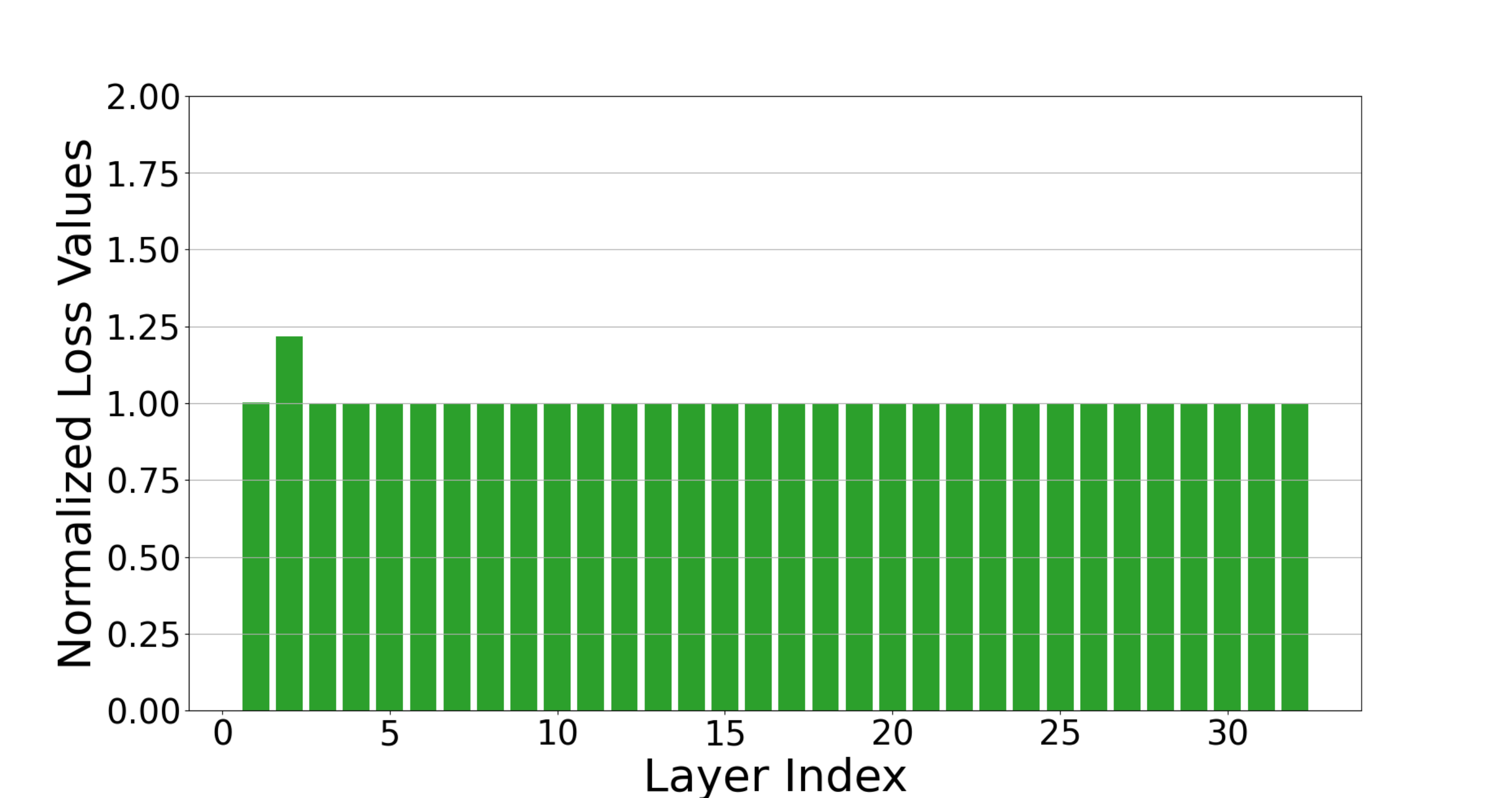}
    \end{minipage}
    \hfill
    \begin{minipage}[b]{0.32\textwidth}
    \includegraphics[width=\textwidth]{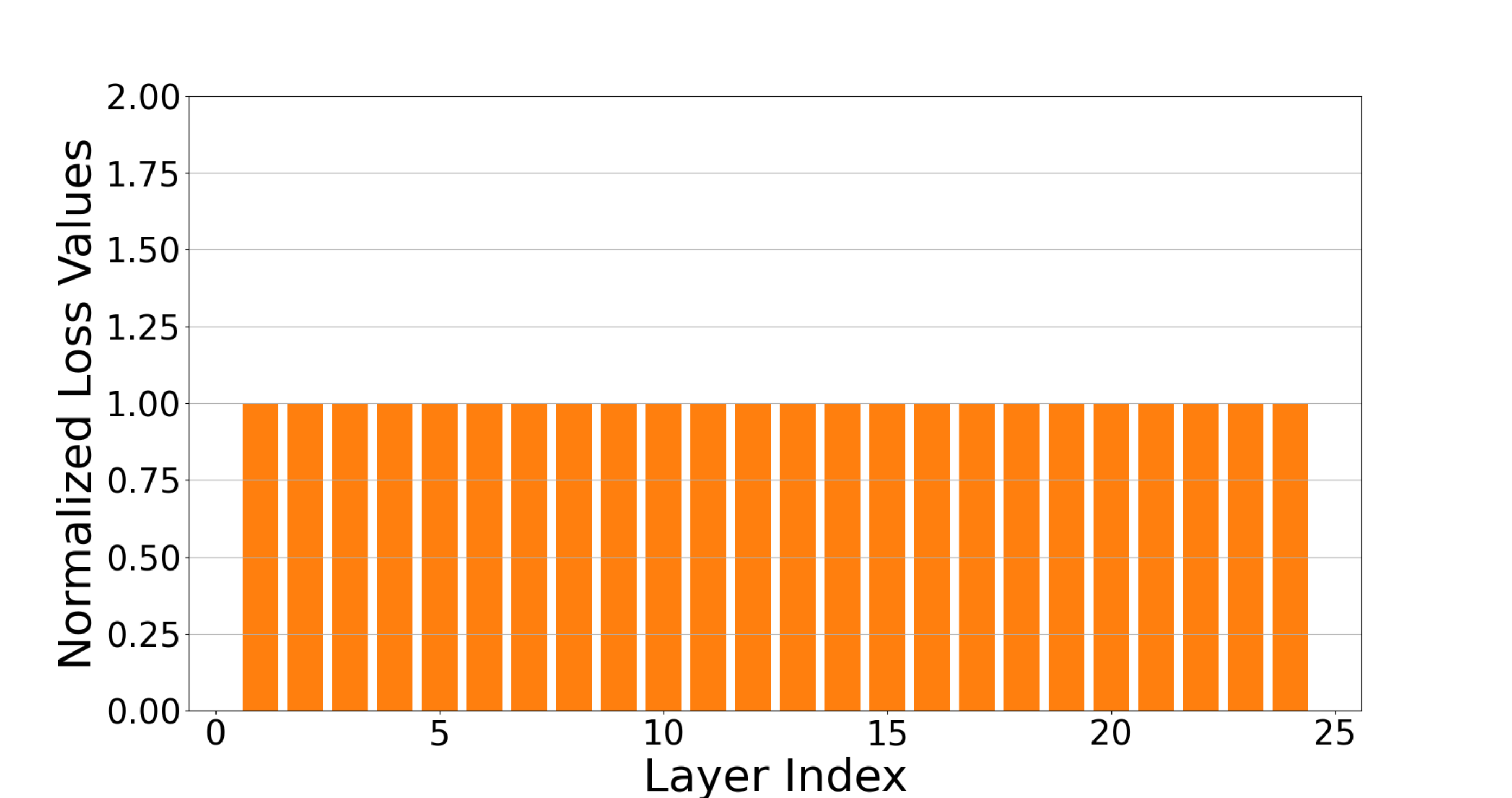}
    \end{minipage}
    \hfill
    \begin{minipage}[b]{0.32\textwidth}
    \includegraphics[width=\textwidth]{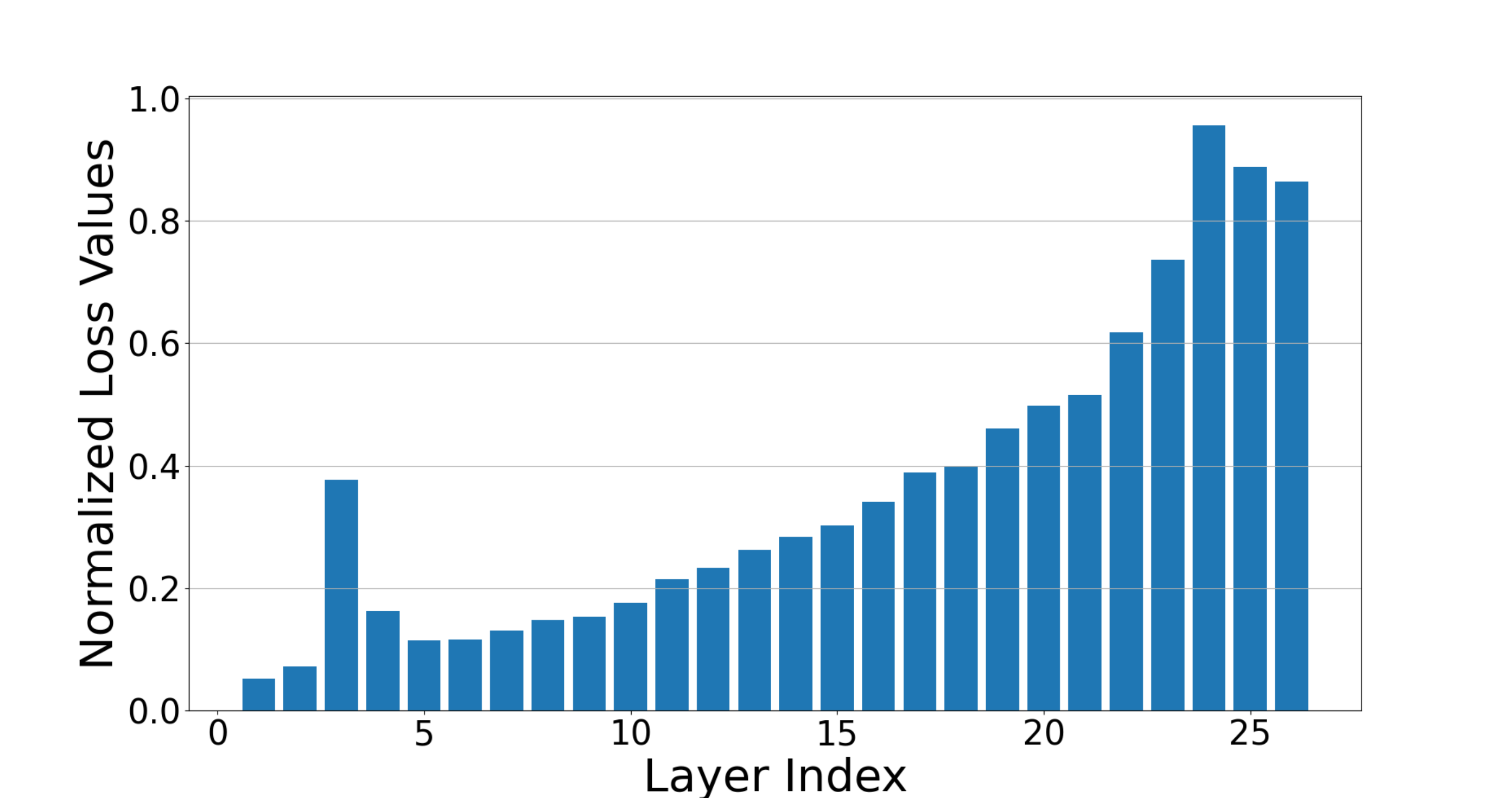}
    \end{minipage}
    \caption{Importance analysis of Mixtral-8$\times$7B (left), Qwen1.5-MoE-A2.7B (middle), and DeepSeek-V2-Lite (right) models. A larger loss indicates greater importance. For Mixtral-8$\times$7B and Qwen1.5-MoE-A2.7B, the importance of the different layers is relatively consistent, but for DeepSeek-V2-Lite, the importance increases as one approaches the output layer.}
    \label{fig:layer}
\vspace{2mm}
\end{figure*}

Following this paradigm, we demonstrate the layer importance for Mixtral-8$\times$7B ~\cite{Mixtral}, Qwen1.5-MoE-A2.7B ~\cite{Qwen}, and DeepSeek-V2-Lite ~\cite{DeekSeekv2} as shown in Figure~\ref{fig:layer}. Note that the previous work ~\cite{EEP} overlooks the varying importance of layers and simply applies a uniform pruning ratio to each layer, leading to a suboptimal solution. In contrast, our analysis shows that some models perform in ways that largely diverge from this strategy. For example, the analysis of DeepSeek-V2-Lite (Figure~\ref{fig:layer}) reveals that layer importance rapidly increases with depth, indicating that deeper layers are more sensitive than shallower ones.

\subsubsection{Inter-Expert Pruning Strategy}
\label{sec:inter-pruning-strategy}
To answer the second question, it is required to identify a combination of $N$ experts that have the least impact on prediction loss. Previous work ~\cite{EEP} utilizes brute-force search to find the least impactful combination of $N$ experts within each layer. However, this method presents two significant drawbacks. First, the brute-force search has high time complexity, making it \textit{extremely time-consuming}, especially when pruning the MoE with a large number of experts. For example, Qwen1.5-MoE-A2.7B and DeepSeek-V2-Lite have 60 and 64 experts per layer, respectively. If 25\% of experts need to be pruned, ~\cite{EEP} needs to traverse $C_{15}^{60}$ and $C_{16}^{64}$ times for each layer respectively, which is unacceptable in terms of time consumption. Second, it restricts the search space within the current layer, only achieving a \textit{local optimum} and potentially missing a more globally optimal solution. 

To mitigate these challenges, we leverage \textit{Genetic Search}~\cite{ga1993, ga2020} with \textit{KT-Receptive Field} methods to enhance search efficiency and concurrently identify the least impactful combinations of experts on a more global scale.


\noindent \textbf{Layer-wise Genetic Search.} To avoid extreme time consumption caused by brute-force search~\cite{EEP}, we leverage the genetic search to select the $M$ candidate combinations in each layer. 

For the $i$-th layer, we define all possible pruning combinations as ${C}_{P_i}$. Here, $P_i$ represents the number of experts to be be pruned in the $i$-th layer. Given that there are $M_i$ experts in the $i$-th layer, ${C}_{P_i}$ denotes the number of combinations for selecting $P_i$ experts to prune from the total of $M_i$ experts.

In the initial stage of Genetic Search, we first initialize a population $\{C_{{P_i},1}, C_{{P_i},2}, \ldots, C_{{P_i},N}\}$, where the population size $N=100$. We then calculate the loss for each combination in the population:
\vspace{1mm}
\begin{equation}
    \mathcal{L}_i^n = \sum_{B} \| \mathcal{F}_i(\mathcal{X}) - \mathcal{F}_i(\mathcal{X}, \{\mathbf{E}_{i}\} \setminus C_{{P_i},n})) \|_F
\vspace{1mm}
\end{equation}
where $\mathcal{F}_i$ represents the output of layer $i$ of the MoE model, and $\|\cdot\|_F$ donates Frobenius norm.

We select the combinations with the smallest loss from $C_{{P_i},n}$ as parents. Using union and random sampling, we generate offspring combinations. Each individual in the offspring population undergoes some mutations, where a few experts to be pruned are randomly replaced. This process is repeated iteratively in 50 steps and we can obtain the optimal a few combinations of expert pruning as candidate combinations in the $i$-th layer. \\ 

\begin{figure}[t]
\vspace{2mm}
  \includegraphics[width=\columnwidth]{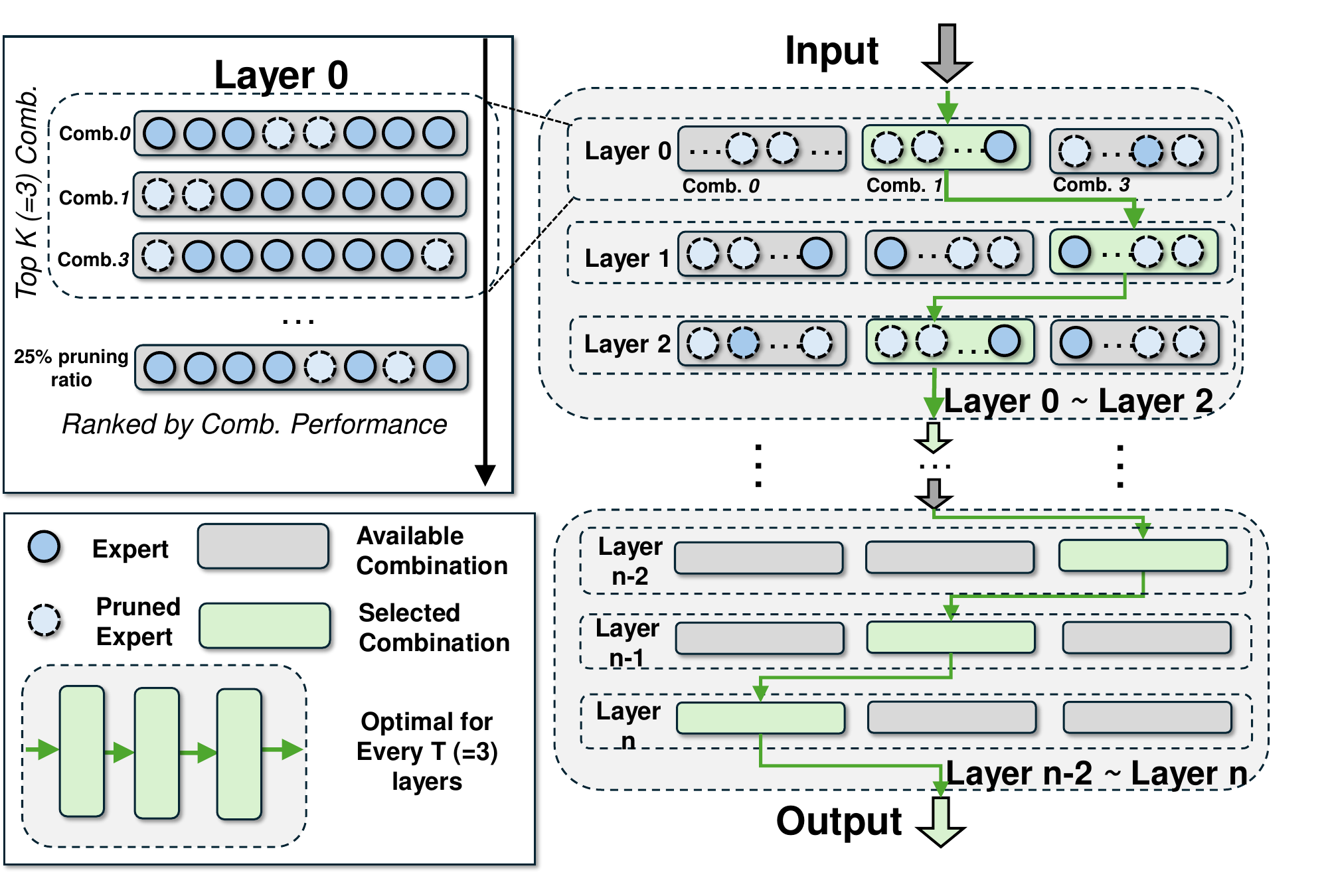}
  \caption{The process of \textit{KT-Receptive Filed} in Mixtral-8$\times$7B for satisfied 25\% pruning ratio. In this case, the number of candidate combinations per layer is $K=3$, and the number of layers per block is $T = 3$. For each layer, we select $K$ optimal candidates using the Layer-wise Genetic Search (top-left). Within a consecutive sequence of $T$ layers, we employ the Block-wise KT-Reception Field to identify the best-performing combination within that block ($T$ layers).} 
  \label{fig:expert_pruning}
\vspace{2mm}
\end{figure}

\noindent \textbf{Block-wise KT-Reception Field.} After obtaining the $n$ candidate combinations, we only keep $K$ best combinations with the smallest loss in each layer as the candidate combinations to be used for the block-level optimization. We aim to select one of the K combinations from each layer such that they minimize the output loss. During this selection process, instead of only considering the importance of experts in just the current layer~\cite{EEP}, we extend the scope of candidate selection from one layer to $T$ layers, achieving a block-wise combination. Specifically, we partition all layers into $\left\lceil \frac{L}{T} \right\rceil$ blocks. Within each block, we select the combination in a brute-force scheme. Given $K$ candidates in each layer, and considering there are $T$ layers in one block, we traverse all possible combinations by selecting one combination from each of $T$ layers, yielding a total of $K^T$ options. Subsequently, we calculate the output loss and select the optimal combinations for pruning. The pipeline is shown in Figure~\ref{fig:expert_pruning}. \\ 

\noindent \textbf{Expert Pruning.} Given the to-be-pruned experts, we conduct the expert pruning operation by removing the entire expert in a structured manner. 

\subsection{Intra-Expert Decomposition}
\label{sec:decomposition}
In this stage, we propose to further compress the remaining experts in a fine-grained way by performing the low-rank decomposition on the parameters within each intra-expert. 

\subsubsection{Expert Importance Analysis}
\label{sec:expert_analysis}
As mentioned in ~\cite{Chi2022}, each expert has varying levels of importance. To achieve better compression performance, instead of applying a uniform compression ratio, we aim to retain more parameters in the important experts and fewer in the less important ones. That leads us to assign higher ranks to the more important experts and lower ranks to the less important ones. Therefore, to calculate the varying ranks, we analyze the relative importance of each expert. Based upon the previous analysis in Sec.~\ref{sec:layer_analysis}, we adopt the same importance metric, $I_{i,j}$ in Eq.~\ref{eq:expert}, as the expert importance. 

To determine the varying ranks of each expert, we begin by calculating basic uniform rank values. Given the overall compression ratio in the second stage, and considering that the structure of all experts is entirely consistent, we directly calculate the target average rank for each expert after decomposition, which is denoted as $\mathcal{R}_a$. By considering the important score of each expert, we calculate the rank values for experts $e_{i,j}$ as:
\vspace{1mm}
\begin{equation}
\mathcal{R}_{ij} = \left\lfloor \frac{(I_{ij} + \epsilon)^\alpha}{\sum_{j=1}^{M_i^{'}} (I_{ij} + \epsilon)^\alpha} \cdot \mathcal{R}_a \cdot M_i^{'} \right\rfloor
\vspace{1mm}
\end{equation}
Here, $M_i^{'}$ represents the number of experts remaining in layer $i$ of the model obtained after \textit{Intra-Expert Pruning}, and $\mathcal{\alpha}$ denotes the smooth factor used to avoid overly linearizing the distribution of rank values, set as $0.15$. $\mathcal{\epsilon}$ is set to $1 \times 10^{-6}$ to avoid the numerical issue.


\subsubsection{Intra-Expert Decomposition Strategy} 
\label{sec:intra-decomposition-strategy}

Singular Value Decomposition (SVD) is a general technique to reduce parameter size by decomposing a large dense matrix into two smaller low-rank matrices. Compared to the Vanilla SVD, which only focuses on the initial weight matrix, ~\cite{SVDLLM2024} generates activation by truncation-aware data whitening and provides hierarchical closed-form updates for model compression. Inspired by SVD-LLM ~\cite{SVDLLM2024} working on dense models, we extend SVD-LLM to MoE models by integrating the non-uniform ranks $R_{i,j}$ in Sec.~\ref{sec:expert_analysis}. 

\subsection{Efficient Fine-tuning}
\label{sec:finetuning}

To mitigate performance degradation caused by the two-stage compression, we fine-tune the MoE by updating the weights. Instead of adjusting all weights, we integrate LoRA ~\cite{lora}, a low-rank approximation technique, into the post-training of the pruned model. The overall algorithm is illustrated in Alg.~\ref{alg:pipeline}.

\renewcommand{\algorithmicrequire}{\textbf{Inputs:}}
\renewcommand{\algorithmicensure}{\textbf{Outputs:}}
\begin{algorithm}
\caption{The Algorithm of MoE-I$^2$}
\small
\begin{algorithmic}[1]
\Require Initial Model $\mathcal{M}$, Target Pruning Ratio $\mathcal{P}_S$, Expert Decomposition Rate $\mathcal{D}$, Calibration Sample $\mathcal{S}_c$, Finetune Sample $\mathcal{S}_f$.
\Ensure Compressed MoE-I$^2$, $\mathcal{M}_f$ 
\For{each layer $l_i$ in $\mathcal{M}$}
    \State $\mathcal{I}_i$ $\gets$ Layer Importance Analysis with $\mathcal{S}_c$ via Sec.~\ref{sec:layer_analysis};
\EndFor

\State $\mathcal{M}_p$ $\gets$ Inter-Expert Pruning($\mathcal{M}$, $\mathcal{S}_c$, $\mathcal{P}_S$, $\mathcal{I}$) via Sec.~\ref{sec:inter-pruning-strategy};

\For{each layer $l_i$ in $\mathcal{M}_p$}
    \State $\mathcal{R}_{i,j}$ $\gets$ Expert Importance Analysis via Sec.~\ref{sec:expert_analysis};
\EndFor

\State $\mathcal{M}_c$ $\gets$ Intra-Expert Decomposition($\mathcal{M}_p$, $\mathcal{S}_c$, $\mathcal{D}$, $\mathcal{R}$) via Sec.~\ref{sec:intra-decomposition-strategy};

\State $\mathcal{M}_f$ $\gets$ Low-Rank Finetune($\mathcal{M}_c$, $\mathcal{S}_f$) via Sec.~\ref{sec:finetuning};

\end{algorithmic}
\label{alg:pipeline}
\end{algorithm}
\vspace{-0mm}

\begin{table*}[ht]
\centering
\small
\caption{Zero-shot performance of three models under our MoE-I$^2$ Framework. The average is calculated among seven classification datasets. ``P'' denotes the Inter-Expert Pruning operation, ``D'' represents the Intra-Expert Decomposition operation, and ``F'' indicates the ``Fine-tuning'' operation based on LoRA. ``Params'' represents the percentage reduction in the number of expert parameters. In the Inter-Expert Pruning stage, we prune 25\% of the experts. During the Intra-Expert Decomposition stage, for the Mixtral-8$\times$7B model, we decompose the remaining experts with an average rank of 2048, further reducing the parameters by approximately 37.5\%. For the Qwen1.5-MoE-A2.7B and DeepSeek-V2-Lite models, we perform decomposition with an average rank of 512, further reducing the parameters by approximately 38.6\%.}
\scalebox{0.93}{
\begin{tabular}{c|c|c|cccccccccc}
\hline
Model& Method   & Params{$\downarrow$} & ARC-c & ARC-e & BoolQ & HellaSwag & OBQA  & RTE   & WinoGrande& Average\\
\hline
\hline
8$\times$7B&baseline& 0     & 57.17 & 84.01 & 85.35 & 64.88     & 35.00 & 70.40 & 75.93     & 67.53\\
8$\times$7B&P    &25\%      & 51.79 & 81.36 & 84.07 & 61.99     & 32.80 & 71.12 & 75.85     & 65.57\\
8$\times$7B&P+F  &25\%      & 56.23 & 82.49 & 86.42 & 64.48     & 36.00 & 72.92 & 74.98     & \textbf{67.65}\\
8$\times$7B&P+D  &51.79\%   & 40.70 & 71.51 & 67.83 & 45.34     & 26.00 & 61.37 & 67.56     &54.33\\
8$\times$7B&MoE-I$^2$ &51.79\% & 52.20 & 78.22 & 82.62 & 61.07     & 34.00 & 72.20 & 71.50     &\textbf{64.55}\\
\hline
\hline
Qwen&   baseline& 0     & 41.89 & 73.11 & 79.76 & 57.90     & 30.40 & 70.04 & 68.67     & 60.25\\
Qwen&   P    &25\%      & 38.57 & 70.37 & 73.30 & 55.84     & 29.80 & 64.98 & 67.25     & 57.16\\
Qwen&   P+F  &25\%      & 45.14 & 75.93 & 78.01 & 57.83     & 32.80 & 71.12 & 68.51     & \textbf{61.33}\\
Qwen&   P+D  &53.98\%   & 37.71 & 65.91 & 71.41 & 49.34     & 29.40 & 64.26 & 67.88     &55.13\\
Qwen&   MoE-I$^2$&53.98\%   & 41.13 & 71.68 & 75.08 & 53.08     & 30.80 & 66.43 & 66.54     &\textbf{57.82}\\
\hline
\hline
DeepSeek&   baseline& 0     & 46.93 & 78.37 & 79.82 & 58.70     & 34.60 & 60.65 & 71.35     & 61.49\\
DeepSeek&   P    &25\%      & 45.31 & 74.62 & 67.95 & 57.38     & 33.20 & 59.93 & 70.01     & 58.34\\
DeepSeek&   P+F  &25\%      & 47.44 & 78.16 & 79.79 & 60.32     & 35.40 & 74.56 & 71.35     & \textbf{63.86}\\
DeepSeek&   P+D  &53.98\%   & 38.48 & 71.42 & 70.09 & 48.15     & 27.80 & 60.65 & 65.98     &54.65\\
DeepSeek&   MoE-I$^2$&53.98\%   & 42.58 & 71.80 & 76.79 & 55.16 & 32.60 & 70.76 & 67.64     &\textbf{59.62}\\
\hline
\end{tabular}
}
\label{tab:moe-pc}
\vspace{-2mm}
\end{table*}

\begin{table*}[ht]
\centering
\small
\caption{Zero-shot performance comparison with EEP~\cite{EEP} and Wanda~\cite{Sun2023}}
\scalebox{0.95}{
\begin{tabular}{c|c|c|c|c|c|c|c|c|c|c}
\hline
Model& Method   & Params{$\downarrow$}     & ARC-c & ARC-e & BoolQ & HellaSwag & OBQA  & RTE   & WinoGrande& Average\\
\hline
\hline
8$\times$7B&EEP&25\%        & 51.62 & 81.94 & 83.64 & 61.60     & 33.00 & 67.87 & 75.37     & 65.01\\
8$\times$7B&P&25\%       & 51.79 & 81.36 & 84.07 & 61.99     & 32.80 & 71.12 & 75.85     & \textbf{65.57}\\
\hline
\hline
8$\times$7B&Wanda&50\%      & 42.06 & 74.16 & 76.64 & 53.16     & 27.00 & 63.90 & 70.96     & 58.27\\
8$\times$7B&EEP  &50\%      & 48.89 & 78.16 & 81.35 & 57.66     & 29.00 & 61.37 & 72.85     & 61.33\\
8$\times$7B&P &50\%      & 48.38 & 78.66 & 81.41 & 58.35     & 27.00 & 64.62 & 74.19     & \textbf{61.80}\\
\hline
\end{tabular}
}
\label{tab:Comparison_Acc}
\vspace{-2mm}
\end{table*}


\begin{table}[ht]
\centering
\small
\caption{Zero-shot performance of experiment results of comparison with EEP and Wanda. ``\textcolor{blue}{$\downarrow$}'' indicates that lower values are better.}
\scalebox{0.95}{
\begin{tabular}{c|c|c|c|c}
\hline
Model & Method & Params{$\downarrow$} & WikiText2 \textcolor{blue}{$\downarrow$} & PTB \textcolor{blue}{$\downarrow$}\\
\hline
\hline
8$\times$7B & baseline & 0\% & 6.24 & 107.24 \\
\hline
\hline
8$\times$7B & EEP & 25\% & 8.16 & 141.1 \\          
8$\times$7B & P & 25\% & \textbf{8.01} & \textbf{133.38} \\
\hline
\hline
8$\times$7B & EEP & 50\% & 11.02 & 207.4 \\            
8$\times$7B & P & 50\% & \textbf{10.1} & \textbf{185.2} \\
\hline
\end{tabular}
}
\label{tab:Comparison_PPL}
\vspace{-6mm}
\end{table}

\begin{table}[ht]
\centering
\small
\caption{Inference speedup performance comparison with EEP~\cite{EEP} and Wanda~\cite{Sun2023} at a compression rate of 50\% . ``\textcolor{blue}{$\downarrow$}'' indicates that lower values are better.}
\scalebox{0.95}{
\begin{tabular}{c|c|c|c|c}
\hline
Model        & Method   & Mem (GB) \textcolor{blue}{$\downarrow$} & Speedup & Average \\ 
\hline
\hline
8×7B        & baseline  & 87.7     & 1.0×    & 67.53   \\
8×7B        & EEP       & 45.78    & 1.20×   & 61.33   \\
8×7B        & Wanda     & 50.01    & 0.91×   & 58.27   \\
8×7B        & MoE-I²    & 43.49    & \textbf{1.28×}   & 64.55   \\
\hline
\hline
Qwen        & baseline  & 26.67    & 1.0×     & 60.25   \\
Qwen        & MoE-I²    & 14.14    & \textbf{1.12×}   & 57.82   \\
\hline
\hline
DeepSeek    & baseline  & 29.26    & 1.0×    & 61.49   \\
DeepSeek    & MoE-I²    & 15.03    & \textbf{1.13× }  & 59.62   \\ 

\hline
\end{tabular}
}
\label{tab:speedup}
\vspace{-6mm}
\end{table}


\begin{table*}[ht]
\centering
\small
\caption{Comparison of zero-shot performance of the MoE-I$^2$ framework and its components. To ensure the same compression ratio and ease of computation as much as possible, when performing the ``D+F'', we set the average rank value of the experts as $\frac{1}{4}$ of expert dimension, which is 352.}
\scalebox{0.95}{
\begin{tabular}{c|c|c|cccccccc}
\hline
Model& Method     & Params {$\downarrow$}   & ARC-c & ARC-e & BoolQ & HellaSwag & OBQA  &RTE   & WinoGrande& Average\\
\hline
\hline
8x7B&   P+F         & 50\%                    & 50.43 & 78.79 & 82.42 & 59.12     & 32.00 & 70.40 & 74.03     & 63.88\\
8x7B&   D+F         &51.35\%                     & 46.08 & 75.34 & 81.41 & 54.02     & 27.80 & 72.20 & 68.27     & 60.73\\
8$\times$7B&MoE-I$^2$   &51.79\%                     & 52.20 & 78.22 & 82.62 & 61.07     & 34.00 & 72.20 & 71.50     &\textbf{64.55}\\
\hline
\hline
Qwen&   P+F       &50\%                       & 41.89 & 69.15 & 75.20 & 53.97     & 30.20 & 64.98 & 62.43     & 56.83\\
Qwen&   D+F       &57.81\%                       & 36.69 & 69.01 & 74.56 & 47.29     & 29.40 & 72.92 & 68.27     &56.88\\
Qwen&   MoE-I$^2$     &53.98\%                       & 41.13 & 71.68 & 75.08 & 53.08     & 30.80 & 66.43 & 66.54     &\textbf{57.82}\\
\hline
\hline
DeepSeek&P+F      &50\%                       & 39.51 & 70.16 & 68.17 & 53.37     & 26.40 & 64.98 & 63.14     & 55.11\\
DeepSeek&D+F      &57.81\%                    & 69.68 & 70.33 & 74.19 & 51.98     & 29.20 & 71.12 & 67.01     & 57.64\\
DeepSeek&MoE-I$^2$&53.98\%                  & 42.58 & 71.80 & 76.79 & 55.16 & 32.60 & 70.76 & 67.64     &\textbf{59.62}\\
\hline
\end{tabular}
}
\label{tab:compare_moe_p_d}
\end{table*}

\begin{figure*}[t]
\vspace{-4mm}
    \centering
    \begin{minipage}[b]{0.49\textwidth}
    \includegraphics[width=\textwidth]{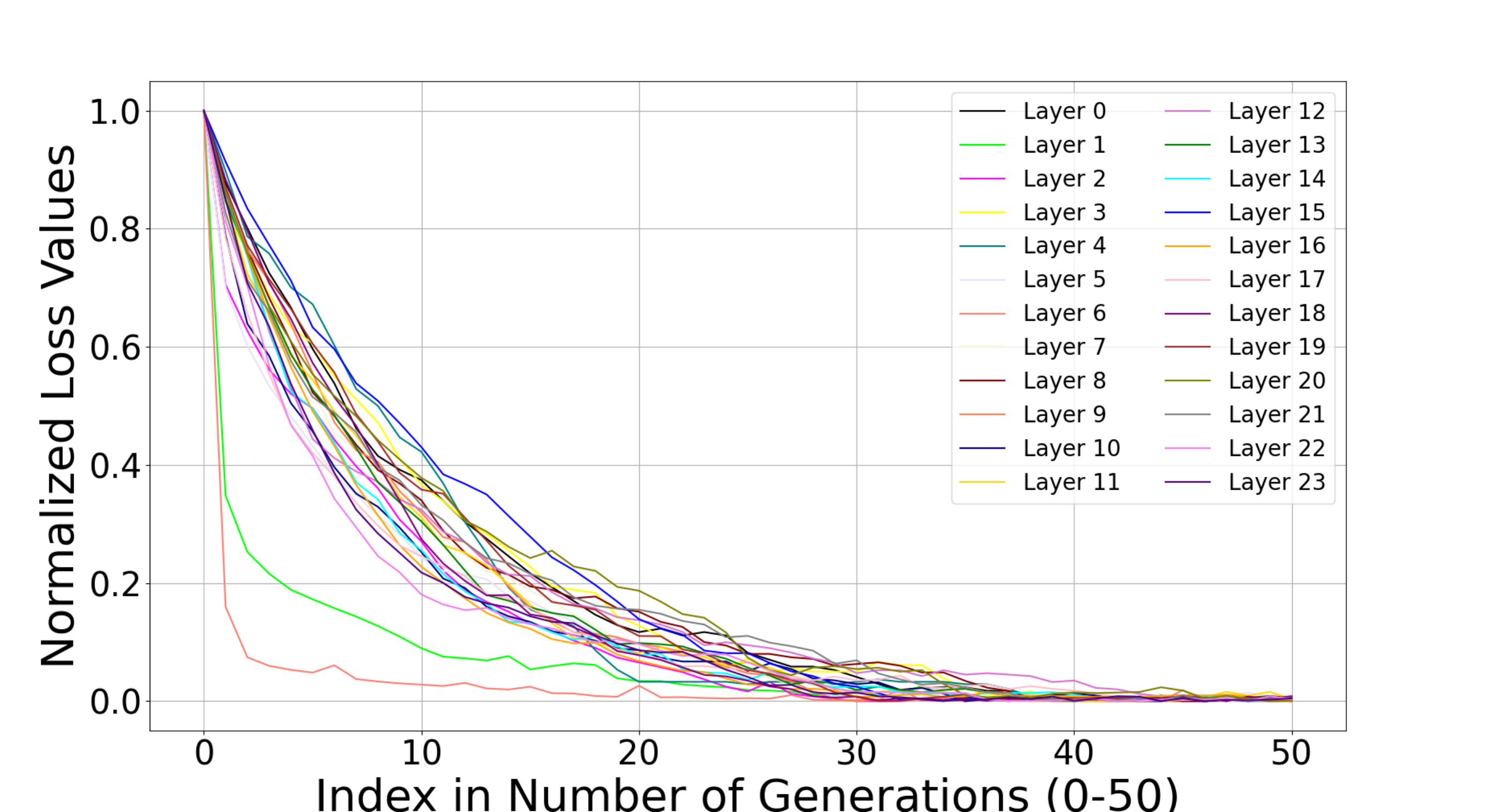}
    \end{minipage}
    \hfill
    \begin{minipage}[b]{0.49\textwidth}
    \includegraphics[width=\textwidth]{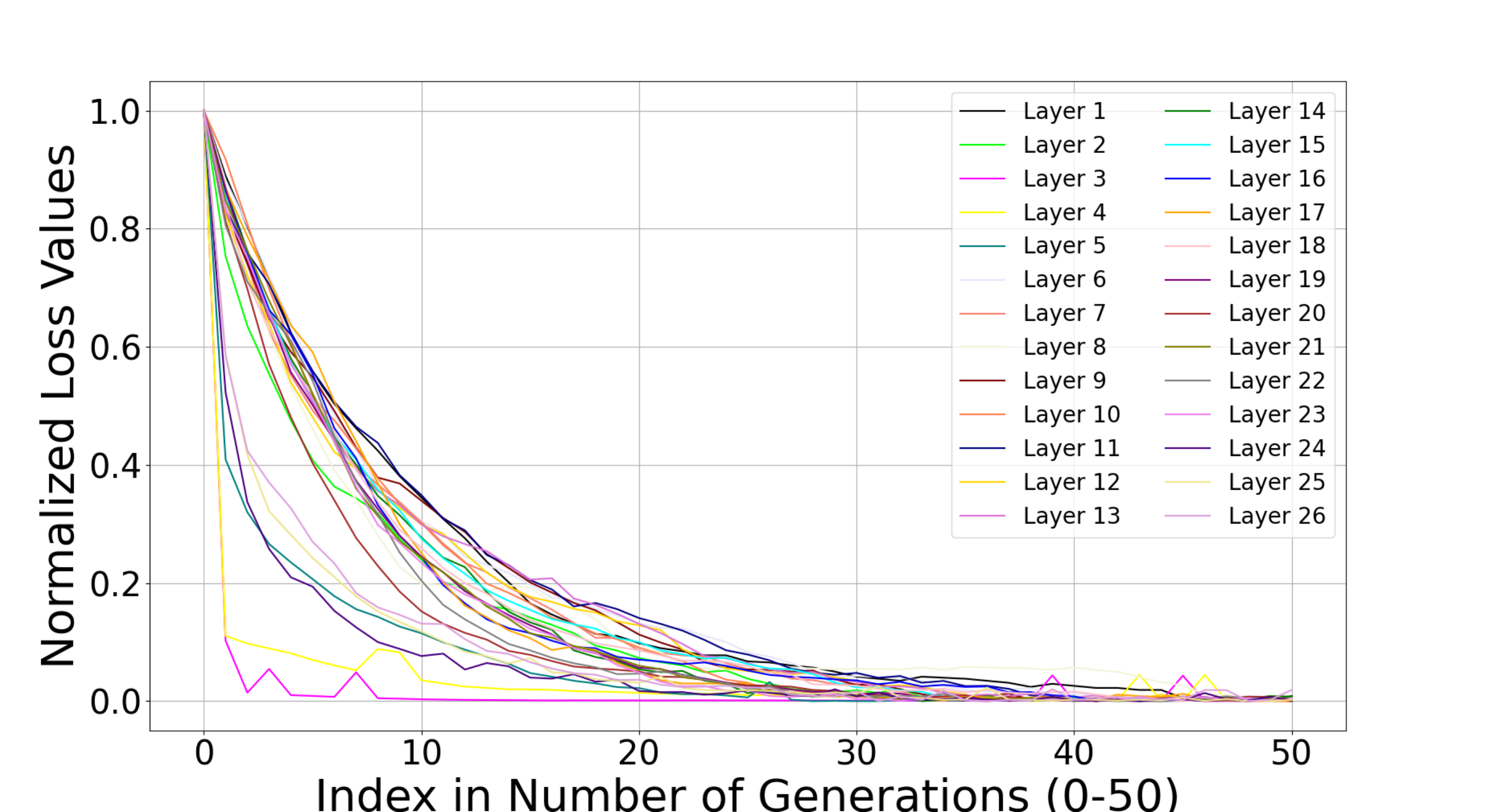}
    \end{minipage}
    \hfill
    \caption{The left and right figures represent the loss convergence for each layer of Qwen1.5-MoE-A2.7B and DeepSeek-V2-Lite during the Genetic Search process, respectively. As shown in the figures, after 50 iterations, nearly all layers have converged.}
    \label{fig:GS}
\vspace{-4mm}
\end{figure*}


\begin{table}[t]
\centering
\small
\caption{Zero-shot performance of average and perplexity of comparison with ``Inter-Expert Pruning'', \textit{Random}, and \textit{TopLoss}. }
\scalebox{0.8}{
\begin{tabular}{c|c|c|c|c|c}
\hline
Model       & Method    & Params{$\downarrow$}  & Average   & WikiText2\textcolor{blue}{$\downarrow$} & PTB\textcolor{blue}{$\downarrow$} \\
\hline
\hline
Qwen        & baseline  & 0\%                   & 60.25     & 7.06      & 13.51 \\
\hline
Qwen        & Random    & 25\%                  & 55.34     & 9.38      & 16.73  \\
Qwen        & TopLoss   & 25\%                  & 56.51     & 8.06      & 15.39 \\
Qwen        & P         & 25\%                  & \textbf{57.16}    & \textbf{8.01}     & \textbf{15.17} \\
\hline
\hline
DeepSeek        & baseline  & 0\%                   & 61.49     & 10.22      & 46.43 \\
\hline
DeepSeek        & Random    & 25\%                  & 43.93     & 48.05      & 628.97  \\
DeepSeek        & TopLoss   & 25\%                  & 57.00     & \textbf{11.34}      & 67.67 \\
DeepSeek        & P         & 25\%                  & \textbf{58.34}    & 11.49     & \textbf{65.80} \\
\hline
\end{tabular}
}
\label{tab:genetic_search}
\vspace{-4mm}
\end{table}

\section{Experiments}
\label{sec:experiments}

\subsection{Experimental Settings}

\noindent\textbf{Model Settings.} To demonstrate the effectiveness of our method, we conducted experiments on three MoE models: Qwen1.5-MoE-A2.7B (14.3B), DeepSeek-V2-Lite (16B), and Mixtral-8$\times$7B (47B). Mixtral-8$\times$7B has a larger number of parameters and relatively fewer experts (8 experts per layer in total 32 layers). On the other hand, Qwen1.5-MoE-A2.7B and DeepSeek-V2-Lite have fewer parameters but a greater number of experts (60 and 64 experts per layer in a total of 24 and 26 layers, respectively). 

\noindent\textbf{Evaluation and Datasets.} To evaluate the performance in a task-agnostic setting, we mainly adopt LLama-Pruner~\cite{llmpruner} evaluation methodology, conducting zero-shot task classification across common sense reasoning datasets such as BoolQ~\cite{boolq}, HellaSwag~\cite{hellaswag}, WinoGrande~\cite{WinoGrande}, ARC-easy~\cite{arc-e}, ARC-challenge~\cite{arc-e}, and OpenbookQA~\cite{openbook}. Meanwhile, our model evaluates results in multiple-choice tasks or generates answers in open-ended generation tasks~\cite{gao2021framework}. Furthermore, we supplement our evaluation with a zero-shot perplexity (PPL) analysis on WikiText2~\cite{wikitext2} and PTB~\cite{PTB}. \\

\noindent\textbf{Implementation Details.}  During the expert pruning phase, we use the same data as the ~\cite{EEP}, which is 2048 randomly sampled data from the C4~\cite{c4} dataset as calibration data. In the expert decomposition phase, we also use 2048 randomly sampled data from Alpaca~\cite{alpaca} as calibration data to conduct the importance analysis. For the finetuning phase, similar to LLM-Pruner~\cite{llmpruner}, we use Alpaca as the finetuning training set, totaling approximately 50k samples. The batch size is set as 64 and learning rates are from 3e-4 to 5e-4. The experiments are conducted on 4 A100-80G GPUs. 

\subsection{Main Results}

\textbf{MoE-I$^2$ Results.} Table~\ref{tab:moe-pc} presents the zero-shot performance of the models after applying the MoE-I$^2$ framework. It is evident that pruning 25\% of the expert parameters results in only a slight performance loss. However, after finetuning the compressed mode with only 2 epochs, the performance can even surpass that of the original model, especially with an improvement of over 2\% on the DeepSeek-V2-Lite model. This observation suggests that pruning 25\% of the experts in the first step is lossless. In the second step, we choose to further compress the pruned model with an approximate 40\% compression ratio via low-rank decomposition. Finally, we perform the finetuning stage. As a result, we can see that while ensuring a reduction of more than 50\% in expert parameters, the model's performance is largely preserved. 

\noindent\textbf{Zero-shot Performance Comparisons with Existing Methods.} 

Table~\ref{tab:Comparison_Acc}  shows the zero-shot performance of the pruned model by comparing Wanda~\cite{Sun2023}, EEP~\cite{EEP}, and our \textit{Inter-Expert Pruning} method under the same sparsity rate. Our method demonstrates significant advantages over Wanda and EEP.

\noindent\textbf{PPL Comparisons with Existing Methods.} Table~\ref{tab:Comparison_PPL}  shows the zero-shot perplexity(PPL) of the pruned model by comparing EEP, and our \textit{Inter-Expert Pruning} method under the same sparsity rate. Our method demonstrates significant advantages over EEP.

\noindent\textbf{Inference Speedup with Existing Methods.} Table~\ref{tab:speedup} shows the speedup of three models by comparing Wanda~\cite{Sun2023}, EEP~\cite{EEP}, and MoE-I$^2$ method.

\subsection{Ablation Studies}
\vspace{1mm}
\noindent\textbf{Comparison of MoE-I$^2$ and its Components.} 
Table~\ref{tab:compare_moe_p_d} demonstrates the necessity of the components within the MoE-I$^2$ framework. It shows that MoE-I$^2$ has a significant advantage when compared to applying only \textit{Inter-Expert Pruning} or \textit{Intra-Expert Decomposition} individually.

\vspace{1mm}
\noindent\textbf{Impact of Genetic Search.} For Qwen1.5-MoE-A2.7B and DeepSeek-V2-Lite models, which have 60 and 64 experts per layer respectively, we only iterate 50 times for Genetic Search. As shown in Figure~\ref{fig:GS}, the loss has converged in the majority of layers. Using EEP~\cite{EEP} for combinatorial search would result in unimaginable time complexity. For instance, if pruning 25\% of the experts, EEP would require searching $C^{60}_{15}$ and $C^{64}_{16}$ times for each layer respectively. Table~\ref{tab:genetic_search} presents the performance of the pruned models obtained through our \textit{Inter-Expert Pruning} compared to \textit{Random} and \textit{TopLoss} methods in terms of zero-shot performance of average(among seven classification datasets) and perplexity tasks. The \textit{TopLoss} denotes that we individually select the $P_i$ least important experts in the current layer to prune instead of considering the expert combination used in Genetic Search. As observed, Genetic Search has a significant advantage over other methods with similar low time costs on seven classification tasks and PPL.

\begin{figure}[t]
\vspace{-4mm}
    \centering
    \includegraphics[width=0.9\linewidth]{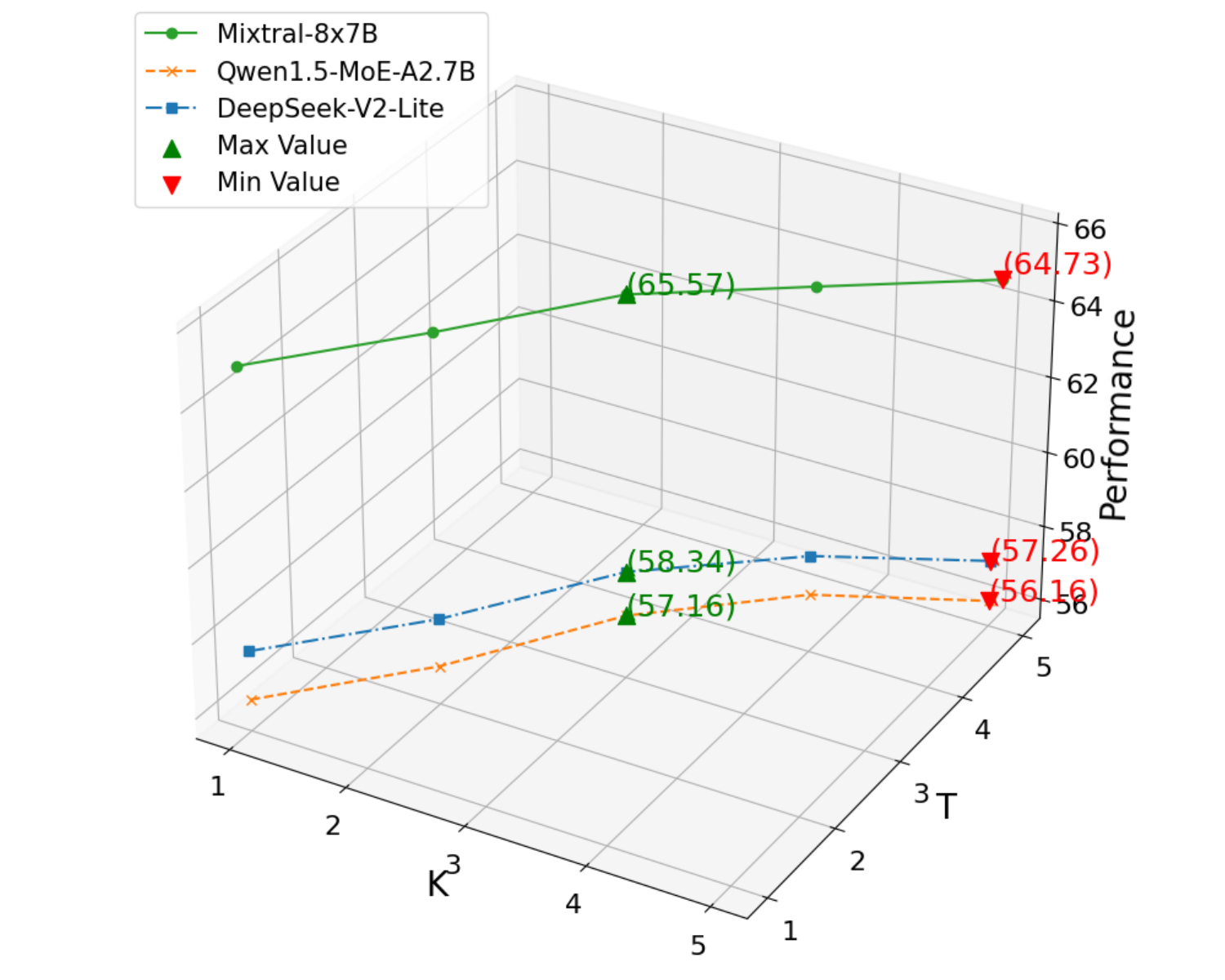}
    \caption{The impact of $K$ and $T$ on the performance of models Mixtral-8$\times$7B, Qwen1.5-MoE-A2.7B, DeepSeek-V2-Lite.}
    \label{fig:KT}
\vspace{-4mm}
\end{figure}

\vspace{1mm}
\noindent\textbf{Impact of KT-Receptive Field.} As shown in Figure~\ref{fig:KT}, we also observe that a large KT-Receptive Field is not always the best during calibration. This is partially because we only use a small amount of data for calibration (2048 samples selected from the C4 dataset). Additionally, there is a significant difference between the C4 dataset and the seven datasets used for zero-shot validation. Simply increasing the values of $K$ and $T$ can lead to overfitting on the calibration dataset. Empirically, $K=3$ and $T=3$ can achieve the best performance. 

\vspace{1mm}
\noindent\textbf{Impact of Non-uniform Pruning Ratio.}  We can observe in Figure~\ref{fig:layer} that the importance of different layers in the DeepSeek-V2-Lite model varies significantly. Table~\ref{tab:deepseek_imbalance} demonstrates that this distinction in layer importance is effective. Compared to the balanced pruning ratio used by Mixtral-8$\times$7B and Qwen1.5-MoE-A2.7B, the imbalance pruning ratio applied to DeepSeek-V2-Lite results in better model performance.

\vspace{1mm}
\noindent\textbf{Impact of Different Ranks.} Table~\ref{tab:expert_diff_rank} shows that selecting an imbalanced rank approach yields better performance for all experts within the same layer. This phenomenon highlights the differences among experts and indicates that different ranks should be assigned to different experts.

\vspace{1mm}
\noindent\textbf{Impact of Experts Pruning, Layers, and Blocks Pruning.} Table~\ref{tab:cmoe} shows our expert pruning method (Genetic Search) demonstrates significant advantages over concurrent approaches, such as Layer Pruning and Block Pruning~\cite{c-moe}. Our Genetic Search can retain more performance (1.32\% vs. 3.19\% performance drop) while maintaining a higher pruning rate (23.95\% vs. 15.51\% pruning ratio). Note that since ~\cite{c-moe} presents normalized zero-shot accuracy results, we have also normalized our results for fairness.


\begin{table*}[ht]
\vspace{-1mm}
\centering
\small
\caption{Zero-shot performance of experiment results produced by Inter-Expert Pruning of comparison with Imbalance ({Imba.}) and Balance ({Ba.}) pruning ratio in DeepSeek-V2-Lite.}
\scalebox{0.95}{
\begin{tabular}{c|c|c|cccccccc}
\hline
Model& Method   & Params{$\downarrow$}     & ARC-c & ARC-e & BoolQ & HellaSwag & OBQA  & RTE   & WinoGrande& Average\\
\hline
\hline
DeepSeek&Ba.    &25\%                      & 44.20 & 73.91 & 68.26 & 57.07     & 32.00 & 57.76 & 69.93     & 57.59\\
DeepSeek&Imba.  &25\%                      & 45.31 & 74.62 & 67.95 & 57.38     & 33.20 & 59.93 & 70.01     & \textbf{58.34}\\
\hline
\hline
DeepSeek&Ba.    &50\%                      & 31.74 & 60.19 & 61.28 & 45.34     & 22.40 & 50.90 & 60.62     & 47.50\\
DeepSeek&Imba.  &50\%                      & 31.74 & 61.87 & 61.74 & 44.79     & 23.60 & 54.87 & 56.67     & \textbf{47.90}\\
\hline
\end{tabular}
}
\label{tab:deepseek_imbalance}
\vspace{-1mm}
\end{table*}


\begin{table*}[ht]
\centering
\small
\caption{Zero-shot performance of experiment results produced by Intra-Expert Decomposition of comparison with Imbalance ({Imba.}) and Balance ({Ba.}) rank in same layer in three models.}
\scalebox{0.95}{
\begin{tabular}{c|c|c|cccccccc}
\hline
Model& Rank(avg) & Type   & ARC-c & ARC-e & BoolQ & HellaSwag & OBQA  & RTE   & WinoGrande& Average\\
\hline
\hline
8x7B&   2048    &Ba.      & 43.66 & 73.45 & 74.03 & 54.31   & 27.40 & 67.92 & 69.55     & 58.62\\
8x7B&   2048    &Imba.    & 43.94 & 73.95 & 74.56 & 55.91   & 27.80 & 68.23 & 69.85     & \textbf{59.18}\\
\hline
8x7B&   1550    &Ba.      & 33.70 & 63.43 & 62.57 & 47.29     & 22.00 & 62.45 & 62.98     & 50.63\\
8x7B&   1550    &Imba.    & 34.59 & 63.67 & 62.59 & 47.68     & 22.00 & 63.05 & 63.15     & \textbf{50.96}\\
\hline
\hline
Qwen&   704     &Ba.      & 40.19 & 72.94 & 77.95 & 54.50     & 30.40 & 68.95 & 69.06     & 59.14\\
Qwen&   704     &Imba.    & 40.44 & 73.40 & 77.74 & 54.54     & 31.60 & 68.95 & 69.30     & \textbf{59.43}\\
\hline
Qwen&   352     &Ba.      & 35.92 & 67.55 & 73.64 & 44.09     & 26.40 & 70.04 & 67.17     & 54.97\\
Qwen&   352     &Imba.    & 36.26 & 67.89 & 73.15 & 44.34     & 27.20 & 72.20 & 66.69     & \textbf{55.39}\\
\hline
\hline
DeepSeek&704    &Ba.      & 43.60 & 76.94 & 77.77 & 53.98     & 30.40 & 62.82 & 69.22     & 59.25\\
DeepSeek&704    &Imba.    & 44.11 & 77.19 & 78.50 & 54.20     & 30.40 & 63.54 & 69.30     & \textbf{59.61}\\
\hline
DeepSeek&352    &Ba.      & 33.45 & 65.11 & 63.05 & 39.07     & 25.20 & 61.75 & 64.88     & 50.35\\
DeepSeek&352    &Imba.    & 34.04 & 65.95 & 63.76 & 39.53     & 25.80 & 60.29 & 65.19     & \textbf{50.65}\\
\hline
\end{tabular}
}
\label{tab:expert_diff_rank}
\vspace{-2mm}
\end{table*}

\begin{table*}[!htbp]
\centering
\small
\caption{Performance of Pruning on Mixtral-8$\times$7B between our Genetic Search and C-MoE~\cite{c-moe}. ``P'' denotes ours Inter-Expert Pruning operation (Genetic Search). ``E[n/m]'' denotes dropping $n$ out of $m$ of experts per MoE layer on average. ``L[n/m]'', ``B[n/m]'' represents dropping $n$ out of $m$ corresponding modules with Layer Drop and Block Drop respectively. These three methods are described in ~\cite{c-moe}.}
\scalebox{0.85}{
\begin{tabular}{c|c|c|c|c|c|c|c|c|c|c}
\hline
Model & Method & Mem(GB) & ARC-c & BoolQ & HellaSwag & OBQA & RTE & WinoGrande & Average & $\Delta$\textcolor{blue}{$\downarrow$} \\
\hline
\hline
8×7B & baseline(Ours/EEP) & 87.7 & 59.81 & 84.92 & 83.97 & 47.00 & 71.12 & 76.32 & 70.52 & - \\
8×7B & P & 66.7 & 56.66 & 83.46 & 81.72 & 46.40 & 71.12 & 75.85 & 69.02 & $\downarrow$ \textbf{1.32} \\
8×7B & baseline~\cite{c-moe} & 87.7 & 59.4 & 84.2 & 84.00 & 46.80 & 70.40 & 75.60 & 70.07 & - \\
8×7B & E2/8 & 66.7 & 53.20 & 77.70 & 80.50 & 46.20 & 55.60 & 76.80 & 65.00 & $\downarrow$ 5.07 \\
8×7B & L8/32 & 66.6 & 47.70 & 85.30 & 75.20 & 40.40 & 69.70 & 74.60 & 65.42 & $\downarrow$ 4.65 \\
8×7B & B5/32 & 74.1 & 51.30 & 85.30 & 78.70 & 42.00 & 69.70 & 74.30 & 66.88 & $\downarrow$ 3.19 \\
\hline
\end{tabular}
}
\label{tab:cmoe}
\vspace{-4mm}
\end{table*}





\section{Conclusion}
\label{sec:conclusion}
In this paper, we explore the efficiency of current large-scale MoE models and propose a general end-to-end compression framework, MoE-I$^2$, that addresses the issue of parameter redundancy in MoE models. In our approach, we first conduct the layer importance analysis and \textit{Inter-Expert Pruning} for different MoE models. Subsequently, we perform the expert important analysis based on the pruned model, ensuring appropriate target ranks of each expert when performing the \textit{Intra-Expert Decomposition}. Our MoE-I$^2$ framework significantly reduces the parameters of MoE models maintaining high performance. In the future, we aim to support a wider variety of MoE models with larger parameters, enhancing their deployability.





\section*{Limitations}
Our proposed framework, MoE-I$^2$, can perform end-to-end compression on any MoE model and adaptively find suitable pruning and decomposition strategies for the target MoE model. By compressing the model at multiple fine-grants, we ensure optimal compression while maintaining model performance, making it more suitable for deployment. Despite these advantages, due to computational limitations, we have not yet tested our framework on larger MoE models such as Mixtral-8$\times$22B (141B), and DeepSeek-V2 (236B). We aim to gradually test these larger MoE models in future work.

\section*{Ethics Statement}
Our research focuses on developing an end-to-end framework for the compression of Mixture-of-Experts (MoE) large language models (LLMs). By enhancing model compression techniques, we aim to significantly reduce the model size and improve inference efficiency, ensuring these improvements do not come at the cost of performance. While our work contributes to the advancement of deploying sophisticated LLMs more effectively, we recognize the ethical considerations inherent in this field. These include the need to address potential biases in the models, ensure the responsible and fair use of LLMs, and safeguard privacy. We are committed to transparency by making our compression framework publicly available. We urge the community to apply our work ethically, with careful attention to the broader societal impacts of deploying compressed LLMs.


\bibliography{custom}






\end{document}